\documentclass[10pt,journal,compsoc]{IEEEtran}
\ifCLASSOPTIONcompsoc
  \usepackage[nocompress]{cite}
\else
  \usepackage{cite}
\fi
\ifCLASSINFOpdf
  \usepackage[pdftex]{graphicx}
  \graphicspath{{figures/}}
\else
  \usepackage[dvips]{graphicx}
  \graphicspath{{figures/}}
\fi
\usepackage{amssymb}
\usepackage{amsmath}
\usepackage{multirow}
\usepackage{array}
\begin{document}
%

\title{Transfer Metric Learning: Algorithms, Applications and Outlooks}

%
%
%
%

\author{Yong~Luo,
        Yonggang~Wen,~\IEEEmembership{Senior~Member,~IEEE,}
        Ling-Yu~Duan,~\IEEEmembership{Member,~IEEE,}
        and~Dacheng~Tao,~\IEEEmembership{Fellow,~IEEE}
\IEEEcompsocitemizethanks{\IEEEcompsocthanksitem Y. Luo and Y. Wen are with the School of Computer Science and Engineering, Nanyang Technological University, Singapore.\protect\\
E-mail: yluo180@gmail.com, ygwen@ntu.edu.sg
\IEEEcompsocthanksitem L. Duan is with the School of Electronics Engineering and Computer Science, Institute of Digital Media, Peking University, China.\protect\\
E-mail: lingyu@pku.edu.cn
\IEEEcompsocthanksitem D. Tao is with the UBTECH Sydney Artificial Intelligence Centre and the School of Information Technologies in the Faculty of Engineering and Information Technologies at University of Sydney, 6 Cleveland St, Darlington, NSW 2008, Australia.\protect\\
E-mail: dacheng.tao@sydney.edu.au
}
}

\markboth{Journal of \LaTeX\ Class Files,~Vol.~XX, No.~X, XXXX~XXXX}%
{Shell \MakeLowercase{\textit{et al.}}: Bare Demo of IEEEtran.cls for Computer Society Journals}

\IEEEtitleabstractindextext{%
\begin{abstract}
Distance metric learning (DML) aims to find an appropriate way to reveal the underlying data relationship. It is critical in many machine learning, pattern recognition and data mining algorithms, and usually require large amount of label information (such as class labels or pair/triplet constraints) to achieve satisfactory performance. However, the label information may be insufficient in real-world applications due to the high-labeling cost, and DML may fail in this case. Transfer metric learning (TML) is able to mitigate this issue for DML in the domain of interest (target domain) by leveraging knowledge/information from other related domains (source domains). Although achieved a certain level of development, TML has limited success in various aspects such as selective transfer, theoretical understanding, handling complex data, big data and extreme cases. In this survey, we present a systematic review of the TML literature. In particular, we group TML into different categories according to different settings and metric transfer strategies, such as direct metric approximation, subspace approximation, distance approximation, and distribution approximation. A summarization and insightful discussion of the various TML approaches and their applications will be presented. Finally, we indicate some challenges and provide possible future directions.
\end{abstract}

}

\maketitle

\IEEEdisplaynontitleabstractindextext

%
\IEEEpeerreviewmaketitle

\IEEEraisesectionheading{\section{Introduction}\label{sec:Introduction}}

%
%
%
%

\IEEEPARstart{I}{t} is critical to evaluate the distances between samples in pattern analysis and machine learning applications. If an appropriate distance metric can be obtained, even the simple $k$-nearest neighbor ($k$-NN) classifier, or $k$-means clustering can perform well \cite{EP-Xing-et-al-NIPS-2002, KQ-Weinberger-et-al-NIPS-2005}. In addition, for large-scale and efficient information retrieval, the results are usually obtained directly according to the distances to the query \cite{P-Jain-et-al-NIPS-2008}, and a good distance metric is also the key of many other important applications, such as face verification \cite{S-Chopra-et-al-CVPR-2005} and person re-identification \cite{LY-Ma-et-al-TIP-2014}.

To learn a reliable distance metric, we usually need large amount of label information, which can be the class labels or target values as used in the typical machine learning approaches (such as classification or regression), and it is more common to utilize some pair or triplet-based constraints \cite{A-Bellet-et-al-arXiv-2014}. Such constraints are weakly-supervised since the exact label for an individual sample is unknown. However, in real-world applications, the label information is often scarce since manually labeling is labor-intensive and it is exhausted or even impossible to collect abundant side information for a new learning problem.

Transfer learning \cite{SJ-Pan-and-Q-Yang-TKDE-2010}, which aims to mitigate the label deficiency issue in model training, is thus introduced to improve the performance of distance metric learnng (DML) when the label information is insufficient in a target domain. This leads to the so-called transfer metric learning (TML), which has been found to be very useful in many applications. For example, in face verification \cite{JL-Hu-et-al-CVPR-2015}, the main step is to estimate the similarities/distances between face images. The data distributions of the images captured under different scenarios vary due to the varied background, illumination, etc. Therefore, the metric learned in one scenario may be not effective in a new scenario and TML would be helpful. In person re-identification \cite{LY-Ma-et-al-TIP-2014, C-Su-et-al-TPAMI-2018}, the key is to estimate the similarities/distances between images of persons appeared in different cameras. The data distributions of the images captured using different cameras vary due to the varied camera setting and scenario. In addition, the distribution for the same camera may change over time. Hence, calibration is needed to achieve satisfactory performance and TML is able to reduce such effort. A more general example is image retrieval, where the data distributions of images in different datasets vary \cite{B-Bhattarai-et-al-CVPR-2016}. It would also be very useful to utilize expensive or semantic features to help learn a metric for cheap features or the ones that are hard to be interpreted \cite{DX-Dai-et-al-CVPR-2015, Y-Luo-et-al-TPAMI-2018}.

In the past decade, dozens of works have been proposed in this area and we provide in this survey a comprehensive overview of these methods. In this survey, we aim to make the machine learners quickly grasp the TML research area, and facilitate the chosen of appropriate methods for machine learning practitioners. Besides, there still be many issues to be tackled in TML, and we hope that some new ideas can be inspired from this survey.

\begin{figure*}[!t]
\centering
\includegraphics[width=1.8\columnwidth]{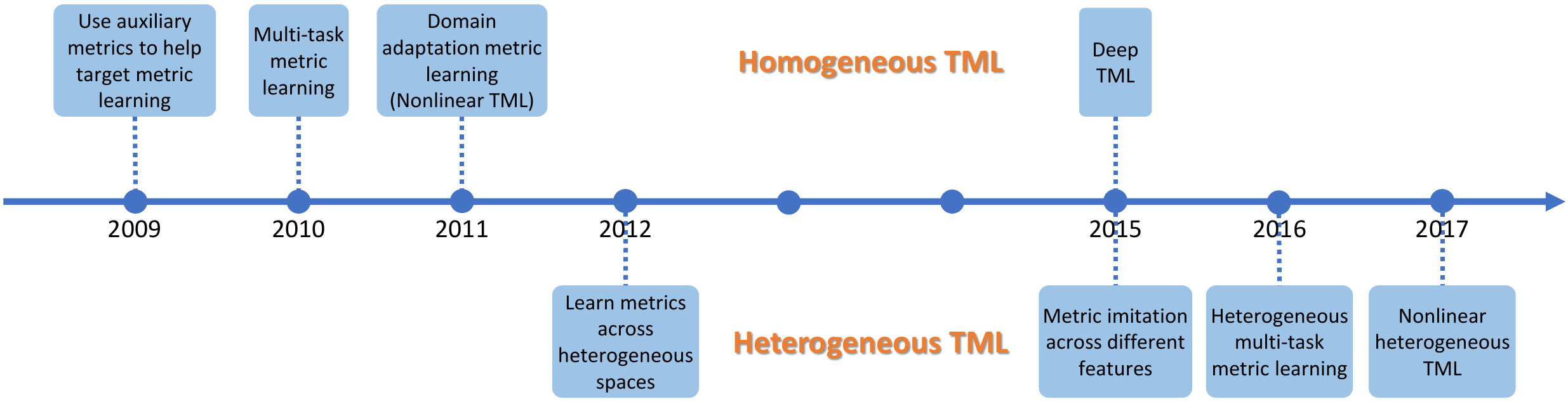}
\caption{Evolution of transfer metric learning, which has been studied for almost ten years.}
\label{fig:TML_Evolution}
\end{figure*}

The rest of this survey is organized as follows. We first present the background and overview of TML in Section \ref{sec:Overview}, which includes a brief history of TML, the main notations used throughout the paper, and a categorization of the TML approaches. In the subsequent two sections, we give a detailed description of the approaches in the two main categories, i.e., homogeneous and heterogeneous TML respectively. Section \ref{sec:Application} is a summarization of the different applications of TML and finally, we conclude this survey and identify some possible future directions in Section \ref{sec:Conclusion}.

%
%

\begin{figure*}[!t]
\centering
\includegraphics[width=1.2\columnwidth]{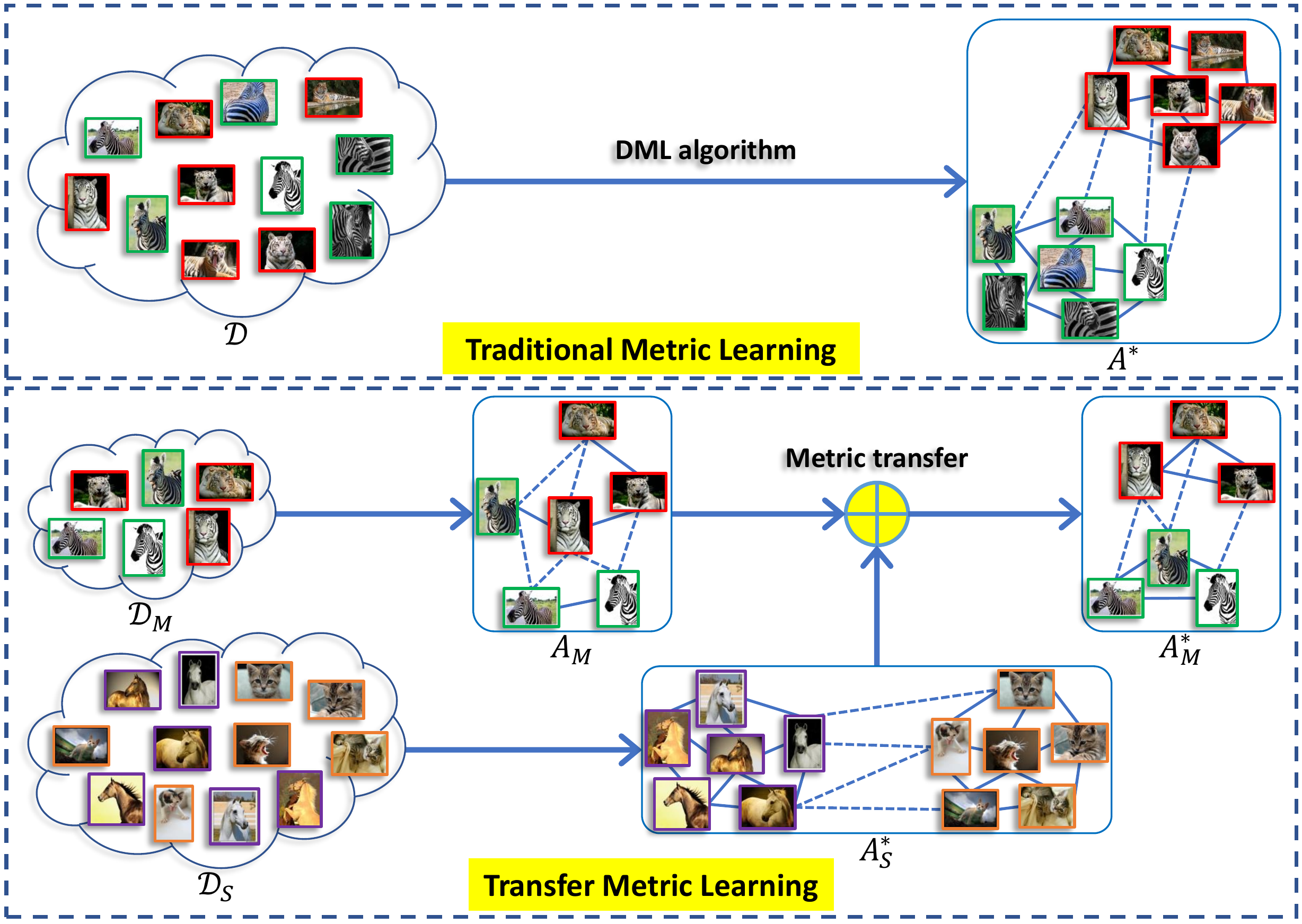}
\caption{An illustration of traditional distance metric learning (DML) and transfer metric learning (TML). Given abundant labeled data, DML aims to learn a distance function between samples so that their distance is small if semantically similar and large otherwise. TML improves DML when the labeled data are insufficient in the target domain by utilizing information from related source domains, which have better distance estimations between samples. For example, it may be hard to distinguish ``zebra'' from ``tiger'' by observing only a few labeled samples due to the very similar stripe texture. But this task can be much easier if we have enough labeled samples to well distinguish ``horse'' from ``cat''. The sample images are from the NUS-WIDE \cite{TS-Chua-et-al-CIVR-2009} dataset.}
\label{fig:DML_vs_TML}
\end{figure*}

\section{Background and Overview}
\label{sec:Overview}

\subsection{A brief history of transfer metric learning}
Transfer metric learning (TML) is a relatively new research field. The works that explicitly applying transfer learning to improve DML start around the year of 2009. For example, multiple auxiliary (source) datasets are utilized in \cite{ZJ-Zha-et-al-IJCAI-2009} to help the metric learning on the target set. The main idea is to enforce the target metric to be close to the different source metrics. An adaptive weight is learned to reflect the contribution of each source metric to the target metric. In \cite{Y-Zhang-and-DY-Yeung-KDD-2010}, such contribution is determined by learning a covariance matrix between the different metrics. Instead of directly learning the target metric, the decomposition based method \cite{Y-Luo-et-al-TIP-2014} assumes that the target metric can be represented as a linear combination of multiple base metrics, which can be derived from the source metric. Hence, the metric learning is casted as learning combination coefficients, where the parameters to be learned can be much fewer.

We can not only using source metrics to help the target metric learning, but also make the different DML tasks help each other. The latter is often called multi-task metric learning (MTML). One representative work is the multi-task extension \cite{S-Parameswaran-and-KQ-Weinberger-NIPS-2010} of a well-known DML algorithm LMNN \cite{KQ-Weinberger-et-al-NIPS-2005}. Some other related works including GPMTML \cite{PP-Yang-et-al-MLJ-2013}, MtMCML \cite{LY-Ma-et-al-TIP-2014} and CP-mtML \cite{B-Bhattarai-et-al-CVPR-2016}. In addition, there are a few domain adaptation metric learning approaches \cite{B-Geng-et-al-TIP-2011, B-Cao-et-al-IJCAI-2011}. Most of the above methods can only learn linear metric for the target domain. The domain adaptation metric learning (DAML) approach presented in \cite{B-Geng-et-al-TIP-2011} is able to learn nonlinear target metric based on the kernel method. Recently, neural network is also employed to conduct nonlinear metric transfer \cite{JL-Hu-et-al-CVPR-2015} by taking the advantage of deep learning technique \cite{Y-Lecun-et-al-Nature-2015}.

The study of heterogeneous TML is a bit later than homogeneous TML and there are much fewer works than those in the homogeneous setting. To the best of our best knowledge, the first work that explicitly designed for heterogeneous TML is the one presented in \cite{GJ-Qi-et-al-SDM-2012}, but it is limited in that only two domains (one source and target domain) can be handled. There exist a few tensor based approaches \cite{Y-Luo-et-al-IJCAI-2016, Y-Luo-et-al-TNNLS-2018} for heterogeneous MTML, where the high-order correlations between all domains are exploited. A main disadvantage of these approaches is that the computational complexity is high. Dai et al. \cite{DX-Dai-et-al-CVPR-2015} proposes an unsupervised heterogeneous TML algorithm, which aims to use some ``expensive'' (sophisticated, off-the-shelf) features to help learn a metric for relatively ``cheap'' feature. This is also termed metric imitation. Recently, a general heterogeneous TML framework is proposed in \cite{Y-Luo-et-al-IJCAI-2017, Y-Luo-et-al-TPAMI-2018}. The framework first extracts some knowledge fragments (linear or nonlinear mappings) from pre-trained source metric, and then using these fragments to help the target domain learn either linear or nonlinear distance metric. The framework is flexible and easy-to-use. An illustration figure for the evolution of TML is shown in Fig.~\ref{fig:TML_Evolution}.

\subsection{Notations and definitions}
In this survey, we assume there are $M$ different domains, and the $m$'th \emph{domain} is associated with a feature space $\mathcal{X}_m$ and marginal distribution $P_m(X_m)$. Without loss of generality, we assume the $M$'th (the last) domain is the target domain, and all the remained ones are source domains. If there is only one source domain, we signify it using the script ``$S$''. In distance metric learning (DML), the \emph{task} is to learn a distance function for any two instances, i.e., $d_\phi(\mathbf{x}_i, \mathbf{x}_j)$, which must satisfy several properties including nonnegativity, identity, symmetry and triangle inequality \cite{A-Bellet-et-al-arXiv-2014}. Here, $\phi$ is the parameter of the distance function, and we call it \emph{distance metric} in this survey. For a nonlinear distance metric, $\phi$ is often given by a nonlinear feature mapping. The linear metric is denoted as $A$, which is a positive semi-definite (PSD) matrix and adopted in the popular Mahalanobias metric learning \cite{EP-Xing-et-al-NIPS-2002}.

To learn the metric in the $m$'th domain, we assume there is a training set $\mathcal{D}_m$, which contains $N_m$ samples with $\mathbf{x}_{mi} \in \mathbb{R}^{d_m}$ to be the feature representation for the $i$'th sample. In a fully-supervised scenario, the corresponding label $y_{mi}$ is also given. However, DML is usually conducted in a weakly-supervised manner, where only some similar/dissimilar constraints on training sample pairs $(\mathbf{x}_{mi}, \mathbf{x}_{mj})$ are provided. Alternatively, the constraint can be a relative comparison for a training triplet $(\mathbf{x}_{mi}, \mathbf{x}_{mj}, \mathbf{x}_{mk})$, e.g., $\mathbf{x}_{mi}$ is more similar to $\mathbf{x}_{mj}$ than to $\mathbf{x}_{mk}$ \cite{A-Bellet-et-al-arXiv-2014}.

In traditional DML, we are often provided with abundant labeled data (such as samples with similar/dissimilar constraints) so that the learned metric $A^\ast$ can well separate semantically similar data from dissimilar ones, such as ``zebra'' and ``tiger'' shown in Fig.~\ref{fig:DML_vs_TML}. While in real-world applications, the learned target metric $A_M$ may be not satisfactory since the labeled data are insufficient in the target domain. For example, it may be hard to distinguish ``zebra'' from ``tiger'' given only a few labeled samples since the two types of animals have very similar stripe texture. To mitigate the label deficiency issue in the target metric learning, we may utilize the information from other related source domain, where the distance metric $A_S^\ast$ is good enough or a good metric can be learned using large amounts of labeled data. For example, if we have enough labeled samples to well distinguish ``horse'' from ``cat'', then it may be very easy for us to recognize ``zebra'' and ``tiger'' by observing only a few labeled samples. The source metric cannot be directly used in the target domain due to the different data distributions \cite{ZJ-Zha-et-al-IJCAI-2009} or representations \cite{GJ-Qi-et-al-SDM-2012} between the source and target domains. Therefore, (homogeneous or heterogeneous) transfer metric learning (TML) is developed to improve the target metric by transferring knowledge (particularly, the metric information) from the source domain. A summarization and discussion of the various TML methods is given as follows.




\subsection{A categorization of transfer metric learning techniques}
\begin{figure*}[!t]
\centering
\includegraphics[width=1.2\columnwidth]{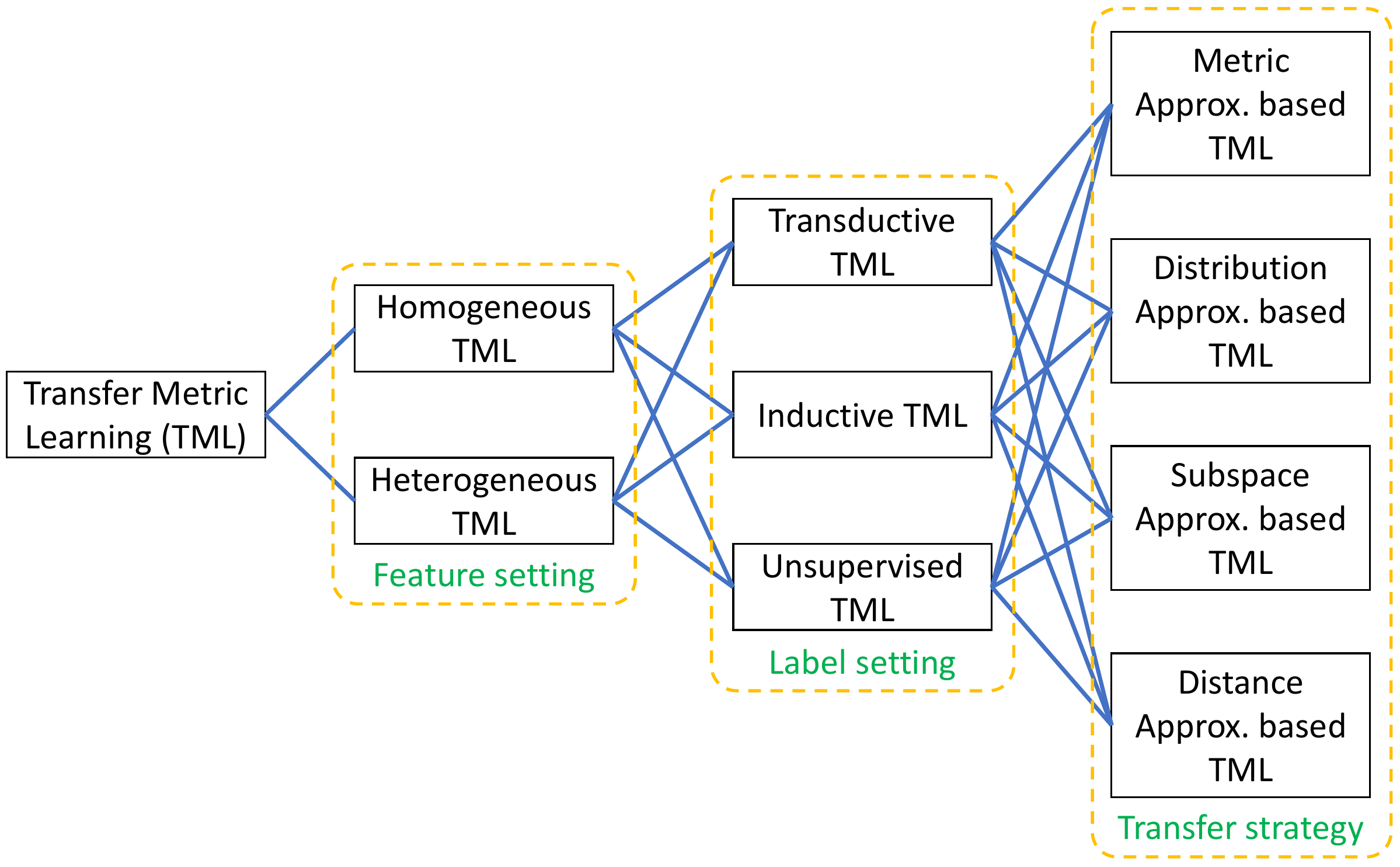}
\caption{A categorization of the TML approaches according to different principals. TML can be categorized according to the feature setting, label setting or utilized transfer strategy. Terms on each path from the left to the right make up a certain TML category, e.g., ``distribution approximation based transductive homogeneous TML''.}
\label{fig:TML_Categories}
\end{figure*}

As shown in Fig.~\ref{fig:TML_Categories}, we can classify TML into different categories according to various principals. Firstly, TML can be generally grouped as \emph{homogeneous TML} and \emph{heterogeneous TML} according to the feature setting. In the former group, the samples of different domains lie in the same feature space $(\mathcal{X}_1 = \mathcal{X}_2 = \cdots = \mathcal{X}_M)$, and only the data distributions vary $(P_1(X_1) \neq P_2(X_2) \neq \cdots \neq P_M(X_M))$. Whereas in heterogeneous TML, the feature spaces are different $(\mathcal{X}_1 \neq \mathcal{X}_2 \neq \cdots \neq \mathcal{X}_M)$ and there may be semantic gap between the source and target domains. For example, in the problem of image matching, we may have only a few labeled images in a new scenario due to the high labeling cost, but there are large amounts of labeled images in some other scenarios. The data distributions of different scenarios vary since there are different backgrounds, illuminations, etc. Besides, the web images are usually associated with text descriptions, and it is useful to utilize the semantic textual features to help learn a better distance metric for visual features \cite{GJ-Qi-et-al-SDM-2012}. The data representations are quite different for the textual and visual domains.

\begin{table*}[!t]
\renewcommand{\arraystretch}{1.3}
\caption{Different categorizes of TML according to label setting.}
\label{tab:TML_Label_Setting}
\centering
\begin{tabular}{|c||c|c|}
  \hline
  TML categorizes & Source domain label information & Target domain label information \\
  \hline \hline
  Inductive TML & Available or Unavailable & Available \\
  \hline
  Transductive TML & Available & Unavailable \\
  \hline
  Unsupervised TML & Unavailable & Unavailable \\
  \hline
\end{tabular}
\end{table*}

\begin{table*}[!t]
\renewcommand{\arraystretch}{1.3}
\caption{Different approaches to TML.}
\label{tab:TML_Strategies}
\centering
\begin{tabular}{|c||p{13.5cm}|}
  \hline
  TML approaches & Brief description \\
  \hline \hline
  Metric approximation & Use the target metric to approximate the source metric \cite{ZJ-Zha-et-al-IJCAI-2009, Y-Zhang-and-DY-Yeung-KDD-2010, Y-Luo-et-al-TIP-2014, HB-Shi-et-al-BMVC-2015, S-Parameswaran-and-KQ-Weinberger-NIPS-2010, PP-Yang-et-al-MLJ-2013}. \\
  \hline
  Distribution approximation & Conduct metric transfer by minimizing the data distributions of different domains \cite{B-Geng-et-al-TIP-2011, JL-Hu-et-al-CVPR-2015, ZM-Ding-and-Y-Fu-TIP-2017, B-Cao-et-al-IJCAI-2011, YH-Xu-et-al-TKDE-2017}. \\
  \hline
  Subspace approximation & Conduct metric transfer by finding a common subspace for different domains \cite{PP-Yang-et-al-NCA-2013, Y-Luo-et-al-IJCAI-2016, Y-Luo-et-al-TNNLS-2018, Y-Luo-et-al-IJCAI-2017, Y-Luo-et-al-TPAMI-2018, GJ-Qi-et-al-SDM-2012}. \\
  \hline
  Distance approximation & Share common parts between distance functions or enforce agreement between distances of corresponding sample pairs in different domains \cite{B-Bhattarai-et-al-CVPR-2016, Y-Luo-et-al-IJCAI-2018, DX-Dai-et-al-CVPR-2015}. \\
  \hline
\end{tabular}
\end{table*}

We can also categorize the different TML approaches as \emph{inductive TML}, \emph{transductive TML}, and \emph{unsupervised TML} according to whether the label information is available in the source or target domains. The relationship of the three learning settings are summarized in Table~\ref{tab:TML_Label_Setting}. This is similar to the categorization of transfer learning presented in \cite{SJ-Pan-and-Q-Yang-TKDE-2010}.

Furthermore, we summarize the TML approaches into four different cases according to the utilized transfer strategies. Some early works of TML directly enforce the target metric to be close the source metric, and we thus refer it to as \emph{TML via metric approximation}. Since the main difference between the source and target domains in homogeneous TML is the distribution divergence, some approaches enable metric transfer by minimizing the distribution difference. We refer this case to as \emph{TML via distribution approximation}. There is a large amount of TML approaches that enable knowledge transfer by finding a common subspace for the source and target domains, especially in heterogeneous TML. This context is referred to as \emph{TML via subspace approximation}. Finally, there is a few works that let the distance functions of different domains share some common parts or enforce the distances of corresponding sample pairs to agree with each other in different domains, and we refer it to as \emph{TML via distance approximation}. The former two cases are usually used in homogeneous TML, and the latter two cases can be adopted for heterogenous TML. Table~\ref{tab:TML_Strategies} is a brief description of these cases.

\begin{table*}[!t]
\renewcommand{\arraystretch}{1.3}
\caption{Different transfer strategies used in different TML settings.}
\label{tab:TML_Strategy_Setting}
\centering
\begin{tabular}{|c||c|c|c||c|c|c|}
  \hline
  \multirow{2}{*}{TML strategies} & \multicolumn{3}{c||}{Homogeneous TML} & \multicolumn{3}{c|}{Heterogeneous TML} \\
  \cline{2-7}
  \ & Inductive & Transductive & Unsupervised & Inductive & Transductive & Unsupervised \\
  \hline
  Metric & $\surd$ & \ & \ & $\times$ & $\times$ & $\times$ \\
  \hline
  Distribution & \ & $\surd$ & \ & \ & \ & \  \\
  \hline
  Subspace & $\surd$ & \ & \ & $\surd$ & \ & $\surd$ \\
  \hline
  Distance & $\surd$ & \ & \ & $\surd$ & \ & $\surd$ \\
  \hline
\end{tabular}
\end{table*}

In Table~\ref{tab:TML_Strategy_Setting}, we show which strategies are currently employed for different settings. In homogeneous TML, most of the current algorithms are inductive, and the transductive ones are usually conducted via distribution approximation. There is still no unsupervised method and a possible solution is to extend some unsupervised DML (e.g., \cite{R-Cinbis-et-al-ICCV-2011}) or transfer learning (e.g., \cite{H-Chang-et-al-TPAMI-2018}) algorithms for unsupervised TML. One challenge is how to ensure the metric learned in the source domain is better since there are no labeled data in both the source and target domains. In the heterogeneous setting \cite{CL-Peng-et-al-TPAMI-2017}, since feature dimensions of different domains do not have correspondences, it is inappropriate to conduct TML via direct metric approximation. Most of the current heterogeneous TML approaches first find a common subspace for different domains, and then conduct knowledge transfer in the subspace. Unsupervised heterogeneous TML can be easily extended for the transductive heterogeneous setting by further utilizing source labels, and it is possible to adopt the distribution approximation strategy in the heterogenous setting by first finding a common representation for the different domains.

\begin{figure*}[!t]
\centering
\includegraphics[width=1.5\columnwidth]{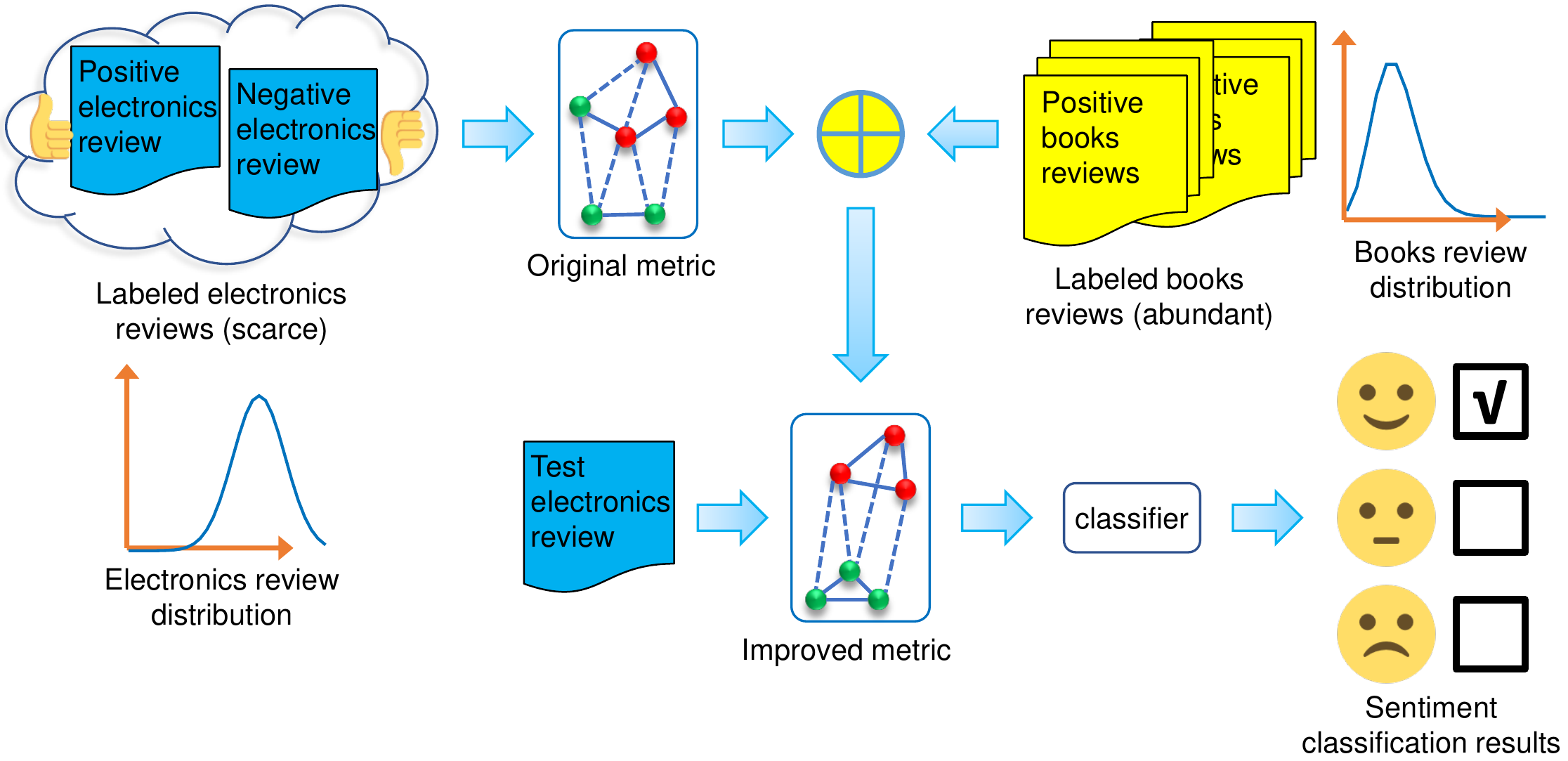}
\caption{An example of homogeneous transfer metric learning. In sentiment classification, distance metric learned for target (such as electronics) reviews may be not satisfactory due to the insufficient labeled data. Homogeneous TML improves the metric by using abundant labeled source (such as book) reviews, where the data distribution is different from the target reviews.}
\label{fig:HoTML_Example}
\end{figure*}

\section{Homogeneous Transfer Metric Learning}
\label{sec:HomoTML}
In homogeneous TML, the utilized features (data representations) are the same, but the data distributions vary for different domains. For example, in sentiment classification as shown in Fig.~\ref{fig:HoTML_Example}, we would like to determine the sentiment polarity (positive, negative or neutral) for a review of electronics. The performance of a sentiment classifier depends much on the distance estimation between reviews. To obtain reliable distance estimation, we usually need large amounts of labeled reviews to learn a good distance metric. However, we may only have a few labeled electronics reviews due to the high labeling cost and thus the obtained metric is not satisfactory. Fortunately, we may have abundant labeled book reviews, which are often easier to collect. Directly applying the metric learned using the labeled book reviews to the sentiment classification of electronics reviews is not appropriate due to the distribution difference between the electronics and book reviews. Transfer metric learning is able to deal with this issue and learn improved distance metric for the target sentiment classification of electronics reviews by using labeled book reviews.

\subsection{Inductive TML}
Under the inductive setting, we are provided with a few labeled data in the target domain. The number of labeled data in the source domain is large enough so that a good distance metric can be obtained, i.e., $N_S \gg N_M > 0$. In inductive transfer learning \cite{SJ-Pan-and-Q-Yang-TKDE-2010}, there may be no labeled source data ($N_S = 0$), but we have not seen such works in homogeneous TML.

\subsubsection{TML via metric approximation}
An intuitive idea for homogeneous TML is to first use the source domain data $\{\mathcal{D}_m\}$ to learn the source distance metrics $\{\phi_m\}$ beforehand, and then enforce the target metric to be close to the pre-trained source metrics. Therefore, the general formulation for learning the target metric $\phi_M$ is given by
\begin{equation}
\label{eq:TML_Met_App}
\mathop{\arg\min}_{\phi_M} \epsilon(\phi_M) = L(\phi_M; \mathcal{D}_M) + \gamma R(\phi_M; \phi_1, \cdots, \phi_{M-1}),
\end{equation}
where $L(\phi_M; \mathcal{D}_M)$ is the empirical loss w.r.t. the metric, $R(\phi_M; \phi_1, \cdots, \phi_{M-1})$ is a regularization term that exploits the relationship between the source and target metrics, and $\gamma \geq 0$ is a trade-off hyper-parameter. Any loss function used in standard DML can be adopted, and the key is how to design an appropriate regularization term. In \cite{ZJ-Zha-et-al-IJCAI-2009}, two different regularization terms are developed. The first one is to minimize the \emph{LogDet} divergence \cite{JV-Davis-et-al-ICML-2007} between the source and target Mahalanobias metrics, i.e.,
\begin{equation}
\label{eq:LDML_LR}
\begin{split}
& R(A_M; A_1, \cdots, A_{M-1}) = \sum_{m=1}^{M-1} \alpha_m D_{LD}(A_M, A_m) \\
& = \sum_{m=1}^{M-1} \alpha_m \left( \mathrm{tr}(A_m^{-1} A_M) - \mathrm{logdet}(A_M) \right).
\end{split}
\end{equation}
Here, $\{A_m \succeq 0\}_{m=1}^M$ are constrained to be PSD matrices and $D_{LD}(\cdot,\cdot)$ indicates the \emph{LogDet} divergence of two matrices. This is more appropriate than the Forbenius norm of matrix difference due to the desirable properties of the \emph{LogDet} divergence, such as scale invariance \cite{JV-Davis-et-al-ICML-2007}. The coefficients $\{\alpha_m\}$ that satisfy $\alpha_m \geq 0$ and $\sum_{m=1}^{M-1} \alpha_m = 1$ is learned to reflect the contributions of different source metrics to the target metric. Secondly, to exploit the geometric structure of data distribution, Zha et al. \cite{ZJ-Zha-et-al-IJCAI-2009} propose a regularization term based on manifold regularization \cite{M-Belkin-et-al-JMLR-2006}:
\begin{equation}
\label{eq:LDML_MR}
R(A_M; A_1, \cdots, A_{M-1}) = \sum_{m=1}^{M-1} \alpha_m \mathrm{tr}\left( X^U L_m (X^U)^T A_M \right),
\end{equation}
where $X^U$ is the feature matrix of unlabeled data, and $L_m$ is the Laplacian matrix of the data adjacency graph calculated based on the metric $A_m$. In \cite{Y-Zhang-and-DY-Yeung-KDD-2010}, the importance of the source metrics to the target metric is exploited by learning a task covariance matrix over the metrics. The matrix can model the correlations between different tasks. This approach allows negative and zero transfer.

Both of the above two approaches incorporate the source metrics into a regularization term to penalize the target metric learning. Different from them, a novel decomposition-based TML method is proposed in \cite{Y-Luo-et-al-TIP-2014}, which constructs the target metric by using the base metrics derived from the source metrics, that is,
\begin{equation}
\label{eq:DTDML_Met}
A_M = U_M \mathrm{diag}(\theta) U_M^T = \sum_{r=1}^{N_B} \theta_{Mr} \mathbf{u}_{Mr} \mathbf{u}_{Mr}^T = \sum_{r=1}^{N_B} \theta_{Mr} B_{Mr},
\end{equation}
where $\{\mathbf{u}_{Mr}\}$ are eigenvectors of source metrics (which are PSD matrices), $\{\theta_{Mr} \geq 0\}$ are combination coefficients of different base metrics, and $N_B$ is the number of bases. This transforms the metric learning into coefficient learning. Hence, the number of parameters to be learned is reduced significantly, and the performance can be improved since the labeled samples in the target domain is scarce. Another advantage of the model is that the PSD constraint of the target metric can be automatically satisfied, and thus the computational cost is low. A semi-supervised extension was presented in \cite{HB-Shi-et-al-BMVC-2015} by combining it with manifold regularization.

In addition to utilizing the source metrics to help the target metric learning, there exist some multi-task metric learning (MTML) approaches that enable different metrics to help each other in metric learning. A representative work is the large margin multi-task metric learning (mtLMNN) \cite{S-Parameswaran-and-KQ-Weinberger-NIPS-2010}, which is a multi-task extension of a well-known DML algorithm, i.e., large margin nearest neighbor (LMNN) \cite{KQ-Weinberger-et-al-NIPS-2005}. In mtLMNN, all the different metrics are learned simultaneously by assuming that each metric consists of a common metric $A_0$ and task-specific metric $\widehat{A}_m$, i.e., $A_m = A_0 + \widehat{A}_m$. Based on the same idea, a semi-supervised MTML method is developed in \cite{Y-Li-and-DC-Tao-ICDMw-2016}, where the unlabeled data is utilized by designing a loss to preserve neighborhood relationship. Then a regularization term is designed to control the amount of information to be shared among all tasks. In \cite{Y-Zhang-and-DY-Yeung-KDD-2010}, a MTML approach is presented by first vectorizing the Mahalanobias metrics and then using a task covariance matrix to exploit the task relationship. Similarly, the metrics are vectorized in \cite{LY-Ma-et-al-TIP-2014}, but the different metrics are enforced to be close under the graph-based regularization theme \cite{T-Evgeniou-et-al-JMLR-2005}. In addition, a general MTML framework is proposed in \cite{PP-Yang-et-al-MLJ-2013}, which enables knowledge transfer by enforcing different metrics $\{A_m\}$ to be close to a common metric $A_0$. The general Bregman matrix divergence \cite{IS-Dhillon-and-JA-Tropp-JMAA-2008} is introduced to measure the difference between two metrics. The framework incorporates mtLMNN as a special case and the geometry is preserved in the transfer by adopting a special Bregman divergence, i.e., the von Neumann divergence \cite{IS-Dhillon-and-JA-Tropp-JMAA-2008}.

\subsubsection{TML via subspace approximation}
Most of the TML approaches via direct metric approximation have a main drawback, i.e., when the feature dimension is high, the model is prone to overfitting due to the large number of parameters to be learned. This also leads to high computational cost in both training and prediction. To tackle this issue, some low-rank TML methods are proposed. They usually decompose the metric as $A_m = U_m U_m^T$, where $U_m \in \mathbb{R}^{d_m \times r}$ is a low-rank transformation matrix. This leads to a common subspace for different domains, and the knowledge transfer is conducted in the subspace. For example, a low-rank multi-task metric learning framework is proposed in \cite{PP-Yang-et-al-ICONIP-2011, PP-Yang-et-al-NCA-2013}, which assumes that each transformation is a product of a common transformation and task-specific one, i.e., $U_m = \widehat{U}_m U_0$. As a special case, the large margin component analysis (LMCA) \cite{L-Torresani-and-K-Lee-NIPS-2007} is extended to multi-task LMCA (mtLMCA), which is shown to be superior to mtLMNN.

\subsubsection{TML via distance approximation}
Both the models of mtLMNN and mtLMCA are trained based on labeled sample triplets. Different from them, CP-mtML \cite{B-Bhattarai-et-al-CVPR-2016} learn the metrics using labeled pairs, which are often easier to collect. Similar to mtLMCA, CP-mtML decomposes the metric as $A_m = U_m U_m^T$, but the different projections $\{U_m\}$ are coupled by assuming that the distance function consists of a common part and task-specific one, i.e.,
\begin{equation}
\label{eq:TML_Dist_App}
d_{U_m}^2(\mathbf{x}_i, \mathbf{x}_j) = d_{\widehat{U}_m}^2(\mathbf{x}_i, \mathbf{x}_j) + d_{U_0}^2(\mathbf{x}_i, \mathbf{x}_j).
\end{equation}
A main advantage of CP-mtML is that the optimization problem can be solved efficiently using stochastic gradient descent (SGD), and hence the model is scalable for high-dimensional features and large amounts of training data. Besides, the learned transformation can be used to derive low-dimensional features, which are desirable in large-scale information retrieval.

\subsection{Transductive TML}
Under the transductive setting, there are no labeled data in the target domain and we only have large amounts of labeled source data, i.e., $N_S \gg N_M = 0$.

\subsubsection{TML via distribution approximation}
In homogeneous TML, the data distributions vary for different domains. Therefore, we can minimize the distribution difference between the source and target domains, so that the source domain samples can be reused in the target metric learning. In \cite{B-Geng-et-al-TIP-2011}, a domain adaptation metric learning (DAML) approach is proposed. In DAML, the distance metric is parameterized by a feature mapping $\phi_M$. The mapping is learned by first transforming the samples in the source and target domains using the mapping, and then minimizing the distribution difference of the source and target domains in the transformed space. At the same time, $\phi_M$ is learned to make the transformed samples satisfy the similar/dissimilar constraints in the source domain. The general formulation for learning $\phi_M$ is given by
\begin{equation}
\label{eq:TML_Distr_App}
\mathop{\arg\min}_{\phi_M} \epsilon(\phi_M) = L(\phi_M; \mathcal{D}_S) + \gamma D_{PD}\left( P_M(X_M), P_S(X_S) \right),
\end{equation}
where $D_{PD}(\cdot,\cdot)$ is a measure of the difference between two probability distributions. Maximum mean discrepancy (MMD) \cite{KM-Borgwardt-et-al-BioInfo-2006} is adopted as the measure in DAML. The nonlinear mapping $\phi_M$ is learned in the reproducing kernel Hilbert space (RKHS), and the solution is found using the kernel method. Since the source and target samples in the transformed space follow similar distribution, the mapping learned using the source label information is also discriminative in the target domain. The same idea is adopted in deep TML (DTML) \cite{JL-Hu-et-al-CVPR-2015}, and the main difference is that the nonlinear mapping is assumed to be a multi-layer neural network. The knowledge transfer is conducted at the output layer and each hidden layer, and some weight hyper-parameters are set to balance the importance of the losses in different layers. A major limitation of these works is that they only consider the marginal distribution difference. This limitation is overcame in \cite{ZM-Ding-and-Y-Fu-TIP-2017}, where a novel TML method is developed by simultaneously reducing the marginal and conditional distribution divergences between the source and target domains. The conditional distribution divergence is reduced by first assigning pseudo labels to target domain data using the classifiers trained on source domain data, and then applying the class-wise MMD \cite{MS-Long-et-al-TKDE-2014}.

Different from these methods, which reduce the distribution difference in a new space, the importance sampling \cite{H-Shimodaira-JSPI-2000} is introduced in \cite{B-Cao-et-al-IJCAI-2011} to handle DML under covariate shift. The formulation is given as follows,
\begin{equation}
\label{eq:CDML}
\mathop{\arg\min}_{A_M \succeq 0} \epsilon(A_M) = \sum_{i,j} w_{ij} l(A_M; \mathbf{x}_{Si}, \mathbf{x}_{Sj}, y_{Sij}),
\end{equation}
where $l(\cdot)$ is some pre-defined loss function over a training pair $(\mathbf{x}_{Si}, \mathbf{x}_{Sj})$ with $y_{Sij} = \pm1$ indicating the two samples are similar or not. The weight $w_{ij} = \frac{P_M(\mathbf{x}_{Si}) P_M(\mathbf{x}_{Sj})}{P_S(\mathbf{x}_{Si}) P_S(\mathbf{x}_{Sj})}$ indicates the importance of the pair in the source domain for learning the target metric. Intuitively, if the pair of source samples have large probability to be occurred in the target domain, they should contribute highly in the target metric learning. In particular, for the distance (such as the popular Mahalanobias distance) which is induced by a norm, i.e., $d(\mathbf{x}_i, \mathbf{x}_j) = \varphi(\mathbf{x}_i - \mathbf{x}_j)$, we can calculate the weight as $w_{ij} = \frac{P_M(\delta_{Sij})}{P_S(\delta_{Sij})}$, where $\delta_{Sij} = \mathbf{x}_{Si} - \mathbf{x}_{Sj}$. In \cite{B-Cao-et-al-IJCAI-2011}, the weights and target metric are learned separately and this may lead to error propagation across them. The issue is tackled by \cite{YH-Xu-et-al-TKDE-2017}, where the weights and target metric are learned simultaneously in a unified framework.

\subsection{Discussion}
TML via metric approximation is straightforward in that divergence between the source and target metrics (parameterized by PSD matrices) are directly minimized. A major difference of the various metric approximation based approaches is that the source and target metrics are enforced to be close in different ways, e.g., by adopting different types of divergence. These approaches are often limited in that the training complexity is high due to the PSD constraint and the distance calculation in the inference stage is not efficient for high-dimensional data. Subspace approximation based TML compensates for these shortcomings by reformulating the metric learning as learning a transformation or mapping. The PSD constraint is automatically satisfied and the learned transformation can be used to derive compressed representation, which would facilitate efficient distance estimation or sample matching, where the hash technique \cite{JD-Wang-et-al-TPAMI-2018} can be involved. This is critical in many applications, such as information retrieval. The main disadvantage of the subspace approximation based methods is that their optimization problems are often non-convex and hence only local optimum can be obtained. The recent work \cite{B-Bhattarai-et-al-CVPR-2016} based on distance approximation also learn a projection instead of the metric but the optimization is more efficient. All of these approaches do not explicitly deal with the distribution difference, which is the main issue that transfer learning would like to tackle. Distribution approximation based methods focus on this point by usually minimizing the MMD measure or utilizing the importance sampling strategy.

\subsection{Related work}
TML is quite related to transfer subspace learning (TSL) \cite{SJ-Pan-et-al-AAAI-2008, S-Si-et-al-TKDE-2010} or transfer feature learning (TFL) \cite{MS-Long-et-al-ICCV-2013}. An early work on TSL is presented in \cite{SJ-Pan-et-al-AAAI-2008} that finds a low-dimensional latent space, where the distribution difference between the source and target domain is minimized. This algorithm is conducted in a transductive manner and not convenient to derive a representation for new samples. This issue is tackled by Si et al. \cite{S-Si-et-al-TKDE-2010}, where a generic regularization framework is proposed for TSL based on Bregman divergence \cite{BA-Frigyik-et-al-TIT-2008}. A low-rank TSL (LTSL) framework is proposed in \cite{M-Shao-et-al-ICDM-2012, M-Shao-et-al-IJCV-2014}, where the subspace is found by reconstructing the projected target data using the projected source data under the low-rank representation \cite{GC-Liu-et-al-ICML-2010, GC-Liu-et-al-TPAMI-2013} theme. The main advantage of the framework is that only relevant source data are utilized to find the subspace and noisy information can be filtered out. That is, it can avoid negative transfer. The framework is further extended in \cite{ZM-Ding-et-al-AAAI-2014} to help recover missing modality in the target domain and improved in \cite{Y-Xu-et-al-TIP-2016} by exploiting both low-rank and sparse structures on the reconstruction matrix.

TFL is very similar to TSL and a representative method is presented in \cite{MS-Long-et-al-ICCV-2013}, where the typical MMD is modified to take both the marginal and class-conditional distributions into consideration. More recent works on TFL are built upon the powerful deep feature learning. For example, considering that the features in deep neural networks are usually general in the first layers and task-specific in higher layers, Long et al. \cite{MS-Long-et-al-ICML-2015} propose the deep adaptation networks (DAN), which frozes the general layers in convolutional neural networks (CNN) \cite{A-Krizhevsky-et-al-NIPS-2012} and only conduct adaption in the task-specific layers. Besides, multi-kernel MMD (MK-MMD) \cite{A-Gretton-et-al-NIPS-2012} is employed to improve kernel selection in MMD. In DAN, only the marginal distribution difference between the source and target domains is exploited. This is improved by the joint adaptation networks (JAN) \cite{MS-Long-et-al-ICML-2017}, which is able to reduce the joint distribution divergence using a proposed joint MMD (JMMD). The JMMD can involve both the input features and output labels in domain adaptation. The constrained deep TSL \cite{Y-Wu-and-Q-Ji-CVPR-2016} method can also exploit the joint distribution and the target domain knowledge is incorporated gradually during a progressive transfer procedure.

\begin{figure*}
\centering
\includegraphics[width=1.5\columnwidth]{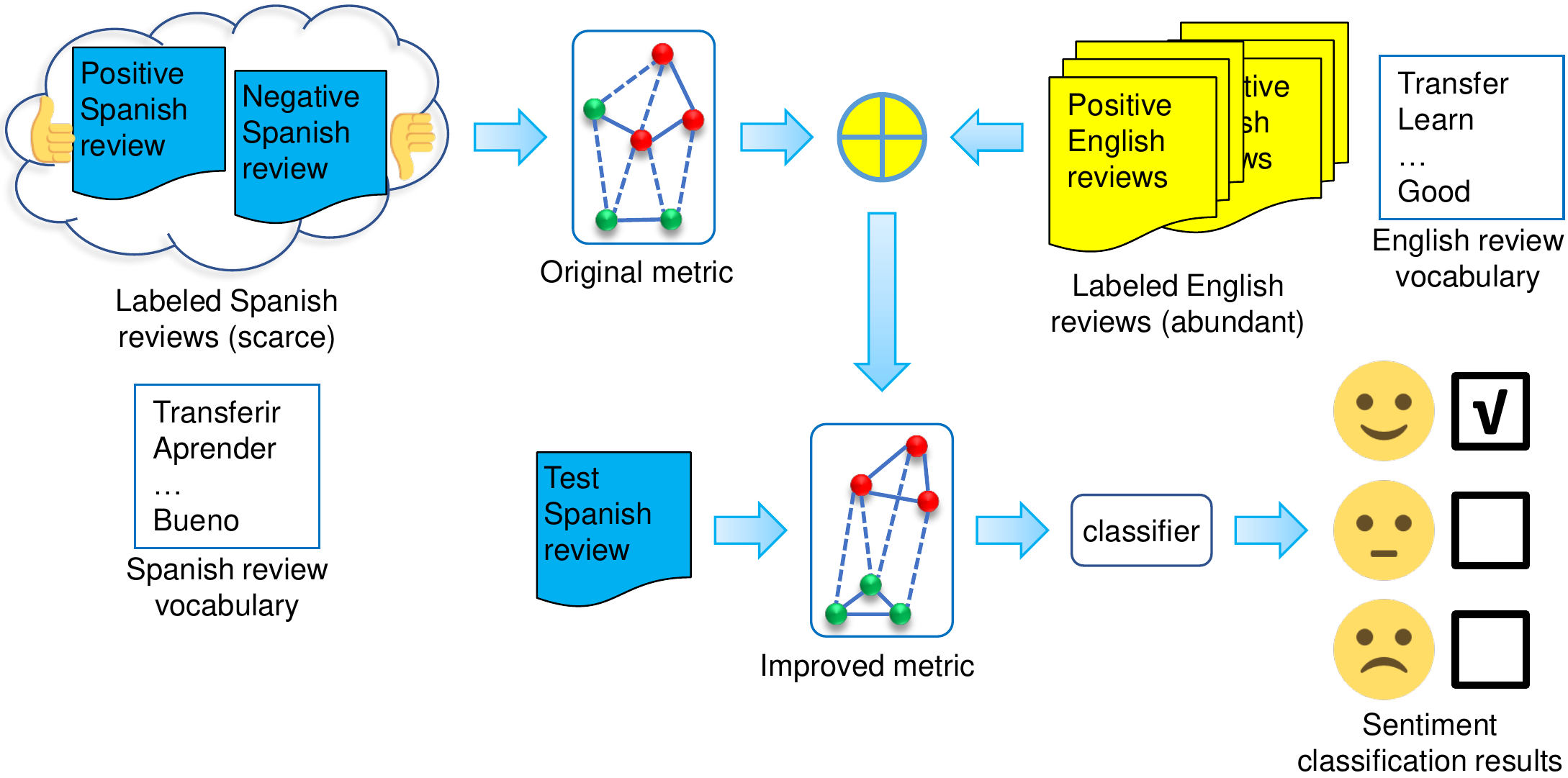}
\caption{An example of heterogeneous transfer metric learning. In multi-lingual sentiment classification, distance metric learned for target reviews (such as the ones written in Spanish) may be not satisfactory due to the insufficient labeled data. Heterogeneous TML improves the metric by using abundant labeled source reviews (such as the ones written in English), where the data representation is different from the target reviews (e.g., due to the different vocabularies).}
\label{fig:HeTML_Example}
\end{figure*}

All of these TSL or TFL approaches have very close relationships to the subspace and distribution approximation based TML. Although they do not aim to learn metrics, it is not hard to adapt them for TML by adopting some metric learning loss in these models.

\section{Heterogeneous Transfer Metric Learning}
\label{sec:HeteTML}
In heterogeneous TML, the different domains have different features (data representations), and sometimes have semantic gap, such as the textual and visual domains. A typical example is the multi-lingual sentiment classification as shown in Fig.~\ref{fig:HeTML_Example}, where we would like to determine the sentiment polarity for a review written in Spanish. The labeled Spanish reviews may be scarce but it is much easier to collect abundant labeled reviews written in English. Directly applying the metric learned using the labeled English reviews to the sentiment classification of Spanish reviews is infeasible since the representations of Spanish and English reviews are different due to the varied vocabularies. This issue can be tackled by heterogeneous TML, which improves the distance metric for the target sentiment classification of Spanish reviews using labeled English reviews.

\subsection{Inductive heterogeneous TML}
Different from the inductive homogenous setting, the number of labeled data in the source domain can be zero under the inductive heterogeneous setting. This is because the source feature may have much stronger representation power than that in the target domain, and thus no labeled data are required to obtain a good distance function in the source domain.

\subsubsection{Heterogeneous TML via subspace approximation}
To our best knowledge, heterogeneous TML under the inductive setting is only studied in recent years. For example, a heterogeneous multi-task metric learning (HMTML) method is proposed in \cite{Y-Luo-et-al-IJCAI-2016}. HMTML assumes that the similar/dissimilar constraints are limited in multiple heterogeneous domains, but there are large amounts of unlabeled data that have representations in all domains, i.e., $\mathcal{D}^U = \{(\mathbf{x}_{1n}^U, \cdots, \mathbf{x}_{Mn}^U)\}_{n=1}^{N^U}$. To build a connection between different domains, the linear metrics $\{A_m\}$ are decomposed as $\{A_m = U_m U_m^T\}$, and then the different representations of an unlabeled data are transformed into a common subspace using $\{U_m\}$. The general formulation is given by
\begin{equation}
\label{eq:HTML_Sub_App}
\mathop{\arg\min}_{\{A_m \succeq 0\}} \epsilon(\{A_m\}) = \sum_{m=1}^M L(A_m; \mathcal{D}_m) + \gamma R(U_1, \cdots, U_M; \mathcal{D}^U).
\end{equation}
Since the different representations corresponding to the same (unlabeled) sample, the transformed representations should be close to each other in the subspace. By minimizing the divergence of transformed representations (or equivalently maximizing their correlations), each transformation is learned by using the information from all domains. This results in an improved transformation, and thus better metric than learning them separately. In \cite{Y-Luo-et-al-IJCAI-2016}, a tensor-based regularization term is designed to exploit the high-order correlations between different domains. A variant of the model is presented in \cite{Y-Luo-et-al-TNNLS-2018}, which uses the class labels to build domain connection.

In \cite{Y-Luo-et-al-IJCAI-2017}, a general heterogeneous TML approach is proposed based on the knowledge fragments transfer \cite{V-Vapnik-and-R-Izmailov-JMLR-2015} strategy. The optimization problem is given by
\begin{equation}
\label{eq:GBHTDML}
\mathop{\arg\min}_{\phi_M} \epsilon(\phi_M) = L(\phi_M; \mathcal{D}_M) + \gamma R(\{\phi_{Mc}(\cdot)\}, \{\varphi_{Sc}(\cdot)\}; \mathcal{D}^U),
\end{equation}
where $\phi_{Mc}(\cdot)$ is the $c$'th coordinate of the mapping $\phi_M$, and $\varphi_{Sc}(\cdot)$ is the $c$'th fragment of the knowledge in the source domain. The source knowledge fragments are represented by some mapping functions, which are learned by applying existing DML algorithms in the source domain beforehand. Then the target metric (which also consists of multiple mapping functions) is enforced to agree with the source fragments on the unlabeled corresponding data. This helps learn an improved metric in the target domain since the pre-trained source distance function is assumed to be superior than the target distance function without knowledge transfer. Intuitively, the target subspace is enforced to approach a better source subspace. An improvement of the model is presented in \cite{Y-Luo-et-al-TPAMI-2018}, where the locality of the geometric structure of the data distribution is preserved via manifold regularization \cite{M-Belkin-et-al-JMLR-2006}.

\subsubsection{Heterogeneous TML via distance approximation}
We can not only enforce the subspace representations of corresponding sample in different domains to be close, but also let the distances of corresponding sample pairs to agree with each other in different domains. For example, an online heterogeneous TML approach is proposed in \cite{Y-Luo-et-al-IJCAI-2018}, which also assumes that there are abundant unlabeled corresponding data, but the target labeled sample pairs are provided in a sequential manner (one by one). Given a new labeled training pair, the target metric is updated as:
\begin{equation}
\label{eq:OHTML}
\begin{split}
& A_M^{k+1} = \mathop{\arg\min}_{A_M \succeq 0} \epsilon(A_M) \\
& = L(A_M) + \gamma_A D_{LD}(A_M, A_M^k) + \gamma_I R(d_{A_M}, d_{A_S}; \mathcal{D}^U),
\end{split}
\end{equation}
where $L(A_M)$ is the empirical loss w.r.t. the current labeled pair, $D_{LD}(\cdot, \cdot)$ is the \emph{LogDet} divergence \cite{JV-Davis-et-al-ICML-2007}, and $R(d_{A_M}, d_{A_S}; \mathcal{D}^U)$ is a regularization term that enforce agreements between the source and target distances (of corresponding pairs). Here, $A_M^k$ is the target metric obtained previously and initialized as an identity matrix. The source metric $A_S$ can be an identity matrix if the source feature is much more powerful than the target feature. By pre-calculating $A_S$ and formulating the term $R(\cdot)$ under the manifold regularization theme \cite{M-Belkin-et-al-JMLR-2006}, an online algorithm is developed to update the target metric $A_M$ efficiently.

\subsection{Unsupervised heterogeneous TML}
There exist a few unsupervised heterogeneous TML approaches that utilize unlabeled corresponding data for metric transfer and no label information is provided in either the source or target domains ($N_S = N_M = 0$). Under this unsupervised paradigm, the utilized source feature should be more expressive or interpretable than the target feature, so that the estimated distances in the source domain can be better than those in the target domain.

\subsubsection{Heterogeneous TML via subspace approximation}
An early work is done in \cite{GJ-Qi-et-al-SDM-2012}, where the main idea is to maximize the similarity of any unlabeled corresponding pairs in a common subspace, i.e.,
\begin{equation}
\label{eq:DT}
\mathop{\arg\min}_{A_M \succeq 0} \epsilon(A_M) = \sum_{n=1}^{N^U} l\left( \varphi(\theta) \right),
\end{equation}
where $\varphi(\theta) = \frac{1}{1 + \exp(-\theta)}$ with $\theta = (\mathbf{x}_{Mn}^U)^T G \mathbf{x}_{Sn}^U$ and $G = U_M^T U_S$. Here, $l(\cdot)$ is chosen to be the negative logistic loss and the proximal gradient method is adopted for optimization. A main disadvantage of this TML approach is that the computational complexity is high since the costly singular value decomposition (SVD) is involved in each iteration of the optimization.

\subsubsection{Heterogeneous TML via distance approximation}
Instead of directly maximizing the likelihood between unlabeled sample pairs, Dai et al. \cite{DX-Dai-et-al-CVPR-2015} propose to use the target samples to approximate the source manifold. The method is inspired by locally linear embedding (LLE) \cite{ST-Roweis-and-LK-Saul-Science-2000}, and metric transfer is conducted by enforcing embeddings of target samples to preserve local properties in the source domain. The optimization problem is given by
\begin{equation}
\label{eq:MI}
\mathop{\arg\min}_{U_M} \epsilon(U_M) = \sum_{i=1}^{N^U} \left\| U_M^T \mathbf{x}_{Mi}^U - \sum_{j=1}^{N^U} w_{Sij}(U_M^T \mathbf{x}_{Mj}^U) \right\|,
\end{equation}
where $w_{Sij}$ is the weight in the adjacency graph calculated using the source domain feature. This enables the distances (between samples) in the source and target domains to agree with each other on the manifold. The optimization is much more efficient than \cite{GJ-Qi-et-al-SDM-2012} since only a generalized eigenvector problem is needed to be solved.

\subsection{Discussion}
It is nature to conduct heterogeneous TML via subspace approximation since the representations of different domains vary and finding a common representation can facilitate the knowledge transfer. Similar to that in the homogeneous setting, the main drawback is that the optimization problem is usually non-convex. Although this drawback can be remedied by directly learning a PSD matrix, such as using the distance approximation strategy, it is nontrivial to perform efficient distance inference for high-dimensional data and extend the algorithm to learn nonlinear metric. Due to the strong ability and rapid development of deep learning, it may be more promising to learn transformation or mapping than PSD matrix in TML, based on either subspace or distance approximation.

\begin{table*}[!t]
\renewcommand{\arraystretch}{1.3}
\caption{A summarization of the different applications in which TML utilized.}
\label{tab:TML_Applications}
\centering
\begin{tabular}{|c||c|p{8cm}|}
  \hline
  \multirow{3}{*}{Homogeneous TML} & Computer vision & Handwritten letter/digit classification, face recognition/verification, image retrieval. \\
  \cline{2-3}
  \ & Speech recognition & English alphabet recognition, vowel classification. \\
  \cline{2-3}
  \ & Other applications & Social network, customer behavior analysis. \\
  \hline
  \multirow{3}{*}{Heterogeneous TML} & Computer vision & Face recognition/verification, scene categorization, image clustering/retrieval/super-resolution. \\
  \cline{2-3}
  \ & Text analysis & Multilingual text categorization, sentiment classification, email spam detection. \\
  \hline
\end{tabular}
\end{table*}

\subsection{Related work}
Some early heterogeneous transfer learning approaches are not specially designed for DML, but the learned feature transformation or mapping for each domain can be used to derive a metric. For example, in the work of heterogeneous domain adaptation via manifold alignment (DAMA) \cite{C-Wang-and-S-Mahadevan-IJCAI-2011}, the class labels are utilized to align different domains. A mapping function is learned for each domain and all functions are learned together. After being projected into a common subspace, the samples should be close to each other if they belong to the same class and separated otherwise. This is conducted for all samples from either the same domain or different domains. The label information of all different domains can be utilized to learn the shared subspace, and thus better embeddings (representations) can be learned for different domains than learning them separately. In \cite{Y-Zhang-and-DY-Yeung-AAAI-2011}, a multi-task discriminant analysis (MTDA) approach is proposed to deal with heterogeneous feature spaces in different domains. MTDA assumes the linear transformation of the $m$'th domain is given by $U_m = W_m H$, which consists of a task-specific part $W_m$ and a common part $H$ for all tasks. Then all the transformations are learned in a single optimization problem, which is similar to that of the well-known linear discriminant analysis (LDA) \cite{K-Fukunaga-Book-1990}. In \cite{X-Jin-et-al-CIKM-2015}, a multi-task nonnegative matrix factorization (MTNMF) approach is proposed to learn the different mappings for all domains by simultaneously factorizing their data representation and feature-class correlation matrices. The factorized class representation matrix is assumed to be shared by all tasks. This leads to a common subspace for different domains.

All of these approaches have very close relationships to the subspace approximation based heterogeneous TML, but they mainly utilize the fully-supervised class labels to learn feature mappings for different domains. As we mentioned previously, it is common to utilize the weakly-supervised pair/triplet constraints in DML and it is not hard to adapt these approaches for heterogeneous TML by adopting some metric learning loss w.r.t. pair/triplet constraints in these models.

\section{Applications}
\label{sec:Application}
In general, for any applications where DML is appropriate, TML is a good candidate when the label information is scarce or hard to collect. In Table~\ref{tab:TML_Applications}, we summarize the different applications that TML utilized in.

\subsection{Homogeneous TML}

\subsubsection{Computer vision}
Similar to DML \cite{B-Kulis-FTML-2012}, most of the TML approaches are applied in computer vision. For example, effectiveness of many homogeneous TML methods are verified in the common image classification application, which includes handwritten letter/digit classification \cite{Y-Zhang-and-DY-Yeung-KDD-2010, PP-Yang-et-al-ICONIP-2011, PP-Yang-et-al-MLJ-2013, Y-Luo-et-al-TIP-2014}, face recognition \cite{ZJ-Zha-et-al-IJCAI-2009, B-Geng-et-al-TIP-2011}, natural scene categorization and object recognition \cite{B-Geng-et-al-TIP-2011, Y-Luo-et-al-TIP-2014, ZM-Ding-and-Y-Fu-TIP-2017}.

DML is particular suitable and crucial for some applications, such as face verification \cite{JL-Hu-et-al-CVPR-2015}, person re-identification \cite{LY-Ma-et-al-TIP-2014} and image retrieval \cite{B-Bhattarai-et-al-CVPR-2016}. This is because in these applications, results can be directly inferred from the distances between samples. Face verification aims to decide whether two face images belong to the same person or not. In \cite{JL-Hu-et-al-CVPR-2015}, TML is applied for face verification across different datasets, where the distributions vary. The goal of person re-identification is to decide whether the people appear in multiple cameras are the same person or not, where the cameras often do not have overlapping views. The data distributions of the images captured by different cameras vary due to the varying illumination, background, etc. Besides, distribution may change over time for the same camera. Hence, TML can be very useful in person re-identification \cite{W-Li-et-al-ACCV-2012, LY-Ma-et-al-TIP-2014, JL-Hu-et-al-CVPR-2015}. An efficient MTML approach is proposed in \cite{B-Bhattarai-et-al-CVPR-2016} to make use of auxiliary datasets for face retrieval, where the tasks vary for different datasets. Stochastic gradient descent (SGD) is adopted for optimization and the algorithm is scalable to large amounts of training data and high dimensional features.

\subsubsection{Speech recognition}
Different groups of speakers have different ways in uttering an English alphabet. In \cite{S-Parameswaran-and-KQ-Weinberger-NIPS-2010, PP-Yang-et-al-MLJ-2013, Y-Li-and-DC-Tao-ICDMw-2016}, alphabet recognition in each group is regarded as a task, and MTML is employed to learn the metrics of different groups together. Similarly, since men and women have different pronunciation styles, vowel classification is performed for two different groups according to the gender, and MTML is adopted to learn their metrics simultaneously by making use of all available labeled data \cite{PP-Yang-et-al-MLJ-2013}.

\subsubsection{Other applications}
In \cite{C-Fang-and-DN-Rockmore-PAKDD-2015}, MTML is used for predictions in social networks. For example, citation prediction is to predict the referencing between articles given their contents. The citation patterns of different areas (such as computer science and engineering) are different but related, and thus MTML is adopted to learn the prediction models of multiple areas simultaneously. Social circle prediction is to assign a person to appropriate social circles given his/her profile. Different types of social circles (such as family members and colleges) are different but related with each other, and hence MTML is applied to improve the performance. In \cite{S-Parameswaran-and-KQ-Weinberger-NIPS-2010, PP-Yang-et-al-MLJ-2013, PP-Yang-et-al-NCA-2013}, MTML is applied to customer information prediction in insurance company. There are multiple variables that can be used to predict the interest of a person in buying a certain insurance policy. Each variable is a discrete value and can be predicted using other variables. The predictions of different variables can be conducted together since they are correlated with each other.

\subsection{Heterogeneous TML}

\subsubsection{Computer vision}
Similar to homogeneous TML, heterogeneous TML is also mainly applied to the computer vision community, such as image classification including face recognition \cite{Y-Zhang-and-DY-Yeung-AAAI-2011}, natural scene categorization \cite{Y-Luo-et-al-TNNLS-2018, Y-Luo-et-al-IJCAI-2017, Y-Luo-et-al-TPAMI-2018} and object recognition \cite{Y-Luo-et-al-IJCAI-2016, Y-Luo-et-al-IJCAI-2017, Y-Luo-et-al-TPAMI-2018}, image clustering \cite{DX-Dai-et-al-CVPR-2015}, image retrieval \cite{DX-Dai-et-al-CVPR-2015, Q-Fu-et-al-TMM-2018, Y-Luo-et-al-TPAMI-2018}, and face verification \cite{Y-Luo-et-al-TPAMI-2018}. In these applications, either the feature dimensions vary or different types of features are extracted for the source and target domains. In particular, expensive features (has strong representation power but high computational cost, such as CNN \cite{K-Chatfield-et-al-arXiv-2014}) can be used to guide learning an improved metric for relatively cheap features (such as LBP \cite{T-Ojala-et-al-TPAMI-2002}), and interpretable text feature can help the metric learning of visual feature, which is often to interpret \cite{GJ-Qi-et-al-SDM-2012, X-Jin-et-al-CIKM-2015}.

In \cite{DX-Dai-et-al-CVPR-2015}, heterogeneous TML is adopted to improve image super-resolution, which is to generate a high-resolution (HR) image for its low-resolution (LR) counterpart. The method is based on JOR \cite{DX-Dai-et-al-CGF-2015}, which is an example-based super-resolution approach. JOR needs to find the nearest neighbors for the LR images, and a metric is learned in \cite{DX-Dai-et-al-CVPR-2015} to replace the Euclidean metric in the $k$-NN search by leveraging information from the HR domain.

\subsubsection{Text analysis}
In the text analysis area, heterogenous TML is mainly applied by using labeled documents written in one language (such as English) to help analysis of the documents in another language (such as Spanish). The utilized vocabularies vary for different languages, and thus the data representations are heterogeneous for different domains. Some typical examples including text categorization \cite{C-Wang-and-S-Mahadevan-IJCAI-2011, Y-Luo-et-al-IJCAI-2016}, sentiment classification \cite{X-Jin-et-al-CIKM-2015} and document retrieval \cite{C-Wang-and-S-Mahadevan-IJCAI-2011}. In \cite{X-Jin-et-al-CIKM-2015}, heterogenous MTML is applied to email spam detection since the vocabularies for different persons' email vary.


\section{Conclusion and Discussion}
\label{sec:Conclusion}

\subsection{Summary}
In this survey, we provide a comprehensive and structured overview of the transfer metric learning (TML) methods and their applications. We generally group TML as homogeneous and heterogeneous TML according to the feature setting. Similar to \cite{SJ-Pan-and-Q-Yang-TKDE-2010}, the TML approaches can also be classified into inductive, transductive and unsupervised TML according to the label setting. According to the transfer strategy, we further categorize the TML approaches into four contexts, i.e., TML via metric approximation, TML via distribution approximation, TML via subspace approximation and TML via distance approximation.

Homogeneous TML has been studied extensively under the inductive setting and various transfer strategies can be adopted. In the transductive setting, TML is mainly conducted by distribution approximation, and there are still no unsupervised methods for homogeneous TML. Unsupervised TML can be carried out under the heterogeneous setting. This is because if more powerful feature is utilized in the source domain, then the distance estimation can be better than that in the target domain \cite{DX-Dai-et-al-CVPR-2015}. Since the data representations vary for different domains in heterogeneous TML, most of these approaches find a common subspace for knowledge transfer.

\subsection{Challenges and future directions}
We finally identify some challenges in TML and speculate several possible future directions.

\subsubsection{Selective transfer in TML}
Current transfer learning and TML algorithms usually assume that the source tasks or domain samples are positively related with the target ones. However, this assumption may not hold in real-world applications \cite{Y-Zhang-and-DY-Yeung-KDD-2010, S-Moon-and-J-Carbonell-IJCAI-2017}. The TML algorithm presented in \cite{Y-Zhang-and-DY-Yeung-KDD-2010} can leverage negatively correlated task by learning task correlation matrix. In \cite{AR-Zamir-et-al-CVPR-2018}, the relations of $26$ popular visual learning tasks are learned using a large image dataset, where each image has annotations in all tasks. This leads to a task taxonomy map, which can be used to guide the chosen of appropriate supervision policies in transfer learning. Different from these approaches, which consider selective transfer \cite{WS-Chu-et-al-TPAMI-2017} at the task-level, a heterogeneous transfer learning method based on the attention mechanism is proposed in \cite{S-Moon-and-J-Carbonell-IJCAI-2017}, which can avoid negative transfer at the instance-level. The low-rank TML model presented in \cite{M-Shao-et-al-IJCV-2014} can also avoid negative transfer to some extent by filtering noisy information in the source domain.

Task correlations have been exploited for metric approximation based TML \cite{Y-Zhang-and-DY-Yeung-KDD-2010}, and the attention scheme can be used for subspace approximation based TML following \cite{S-Moon-and-J-Carbonell-IJCAI-2017}. It is still unclear how to conduct selective transfer in distribution and distance approximation based TML. Adopting the attention scheme may be a certain choice, but this scheme cannot make use of the negative transfer. Therefore, a promising future direction may be to conduct selective transfer at the hypothesis space-level so that both the positive and negative transfer can be effectively utilized.

\subsubsection{More theoretical understanding of TML}
There is a theoretical study in \cite{Y-Luo-et-al-TPAMI-2018}, which shows that generalization ability of the target metric can be improved by directly enforcing the source feature mappings to agree with the target mappings. But there is still lack of general analysis scheme (such as \cite{MM-Gong-et-al-ICML-2016, TL-Liu-et-al-TPAMI-2017, TL-Liu-et-al-IJCAI-2017}) and theoretical results for TML. In particular, more theoretical studies should be conducted to understand when and how could the source domain knowledge help the target metric learning.

\subsubsection{TML for handling complex data}
Most of current TML approaches only learn linear metrics (such as the Mahalanobis metric). However, there may be nonlinear structure in the data, e.g., most of the visual feature representations. Linear metric may fail to capture such structure and hence it is desirable to learn nonlinear metric for the target domain in TML. There have been several works on nonlinear homogeneous TML based on neural networks \cite{JL-Hu-et-al-CVPR-2015, MS-Long-et-al-ICML-2015, MS-Long-et-al-ICML-2017}. But all of them are mainly designed for continuous real-valued data and learn real-valued metrics. More studies can be conducted for histogram data or learning binary target metrics. The histogram data is popular in visual analytic-based applications, and binary metric is efficient in distance calculation. As far as we are concerned, there is only one nonlinear TML work under the heterogeneous setting \cite{Y-Luo-et-al-IJCAI-2017} (with an extension presented in \cite{Y-Luo-et-al-TPAMI-2018}), where gradient boosting regression tree (GBRT) \cite{JH-Friedman-AoS-2001, D-Kedem-et-al-NIPS-2012} is adopted to learn nonlinear metric in the target domain. Some other nonlinear learning techniques can be investigated, and also binary metrics can be learned to accelerate prediction. In addition, when the structure of the target data distribution is very complex, it could be a good choice to learn Riemannian metric \cite{ZW-Huang-et-al-TPAMI-2018} or multiple local metrics \cite{YK-Noh-et-al-TPAMI-2018} to approximate the geodesic distance in the target domain.

\subsubsection{TML for handling changing and big data}
In TML, all the training data in the source and target domains are usually assumed to be provided at once and a fixed target metric is learned. However, in real-world applications, the data are usually comes in a sequential order and the data distribution may change overtime. For example, tremendous amounts of data are uploaded on the web everyday, and for a robot, the environment changes overtime and feedbacks are provided continuously. Therefore, it is desirable to develop some TML algorithms to make the metric adapt to different changes. Some quite related topics including online learning \cite{T-Anderson-Book-2008, B-Wang-et-al-TPAMI-2017} and lifelong learning \cite{S-Thrun-and-L-Pratt-Book-1998}. There is a recent try in \cite{Y-Luo-et-al-IJCAI-2018}, where an online heterogeneous TML is developed. However, this approach needs abundant unlabeled corresponding data in the source and target domains for knowledge transfer. Hence, the approach may be not efficient when vast amounts of unlabeled data are needed to achieve satisfactory accuracy.

Although the number of training data in the target domain is often assumed to be small, the continuously changing data is ``big'' in a long term. In addition, when the feature dimension is high, computational costs of the distances between vast amounts of samples based on a learned Mahalanobias metric is intolerable. A typical example is information retrieval. Therefore, it is desirable to learn some target metric that is efficient in distance calculation, e.g., learn hamming distance metric \cite{M-Norouzi-et-al-NIPS-2012, Z-Wang-et-al-IJCAI-2015} or feature hashing \cite{Z-Wang-et-al-AAAI-2016, LY-Duan-et-al-TMM-2015, LY-Duan-et-al-TIP-2018, JD-Wang-et-al-TPAMI-2018} in the target domain.

\subsubsection{TML for handling extreme cases}
One-shot learning \cite{FF-Li-et-al-TPAMI-2006} and zero-shot learning \cite{M-Palatucci-et-al-NIPS-2009} are two extreme cases of transfer learning. In these cases, the number of labeled data in the target domain is very small (such as only one) and even zero. The main goal is to recognize rare or unseen classes \cite{GJ-Qi-et-al-TPAMI-2017}, where some additional knowledge (such as descriptions of the relations between existing and unseen classes) may be provided. This is more like human learning, and much useful in practice. They are quite related to the concepts of domain generalization \cite{K-Muandet-et-al-ICML-2013, M-Ghifary-et-al-TPAMI-2017, W-Li-et-al-TPAMI-2018}.

DML has been found to be useful in learning unknown classifiers \cite{T-Mensink-et-al-ECCV-2012} (with an extension in \cite{T-Mensink-et-al-TPAMI-2013}), but it does not aim to learn a metric in the target domain. In \cite{C-Fang-et-al-ICCV-2013}, an unbiased metric is learned across different domains, but no specific information about the target domain is leveraged. Although some existing TML algorithms allow no labeled data in the target domain \cite{JL-Hu-et-al-CVPR-2015, DX-Dai-et-al-CVPR-2015}, they need large amounts of unlabeled target data, which can be regarded as additional knowledge. If we do not have unlabeled data, is it possible to utilize other semantic information to help the target metric learning? There exists a try in \cite{S-Bak-and-P-Carr-CVPR-2017}, where the ColorChecker Chart is utilized as additional information for person re-identification under the one-shot setting. But such information is not easy to access and not general for different applications. Hence, more common and easily accessible knowledge should be identified and explored for general TML under the one/zero-shot setting.



%



%
%

\ifCLASSOPTIONcaptionsoff
  \newpage
\fi



\bibliographystyle{IEEEtran}
\bibliography{IEEEabrv,./TML_Survey}
\end{document}